\documentclass[letterpaper]{article} 
\usepackage{aaai19}  
\usepackage{times}  
\usepackage{helvet}  
\usepackage{courier}  
\usepackage{url}  
\usepackage{graphicx}
\usepackage[ruled]{algorithm2e} 
\usepackage{subfig}
\usepackage{amsthm}
\usepackage{algorithmic}
\usepackage{amsmath}
\usepackage{amssymb}
\usepackage{multirow} 
\usepackage[utf8x]{inputenc}
\DeclareUnicodeCharacter{ +65306}{~}
\begin{document}
\title{Similarity Learning via Kernel Preserving Embedding}
\author{Zhao Kang$^{1*}$, 
Yiwei Lu$^{2}$, 
Yuanzhang Su$^3$,
Changsheng Li$^1$,
Zenglin Xu$^1$
\\ 
$^1$School of Computer Science and Engineering, University of Electronic Science and Technology of China, China \\
$^2$  Department of Computer Science, University of Manitoba, Canada
\\
$^3$School of Foreign Languages,
University of Electronic Science and Technology of China, China\\
*Zkang@uestc.edu.cn 
}
\maketitle
\begin{abstract}
Data similarity is a key concept in many data-driven applications. Many algorithms are sensitive to similarity measures. To tackle this fundamental problem, automatically learning of similarity information from data via self-expression has been developed and successfully applied in various models, such as low-rank representation, sparse subspace learning, semi-supervised learning. However, it just tries to reconstruct the original data and some valuable information, e.g., the manifold structure, is largely ignored. In this paper, we argue that it is beneficial to preserve the overall relations when we extract similarity information. Specifically, we propose a novel similarity learning framework by minimizing the reconstruction error of kernel matrices, rather than the reconstruction error of original data adopted by existing work. Taking the clustering task as an example to evaluate our method, we observe considerable improvements compared to other state-of-the-art methods. More importantly, our proposed framework is very general and provides a novel and fundamental building block for many other similarity-based tasks. Besides, our proposed kernel preserving opens up a large number of possibilities to embed high-dimensional data into low-dimensional space.
\end{abstract}

\section{Introduction}
Nowadays, high-dimensional data can be collected everywhere, either by low-cost sensors or from the internet \cite{chen2012fgkm}. Extracting useful information from massive high-dimensional data is critical in different areas like text, images, videos and more. Data similarity is especially important since it is the input for a number of data analysis tasks, such as spectral clustering \cite{ng2002spectral,chen2017dnc}, nearest neighbor classification \cite{weinberger2005distance}, image segmentation \cite{li2016multitask}, person re-identification \cite{hirzer2012relaxed}, image retrieval \cite{hoi2008semi}, dimension reduction \cite{passalis2017dimensionality}, and graph-based semi-supervised learning \cite{kang2018self}. Therefore, similarity measure is crucial to the performance of many techniques and is a fundamental problem in machine learning, pattern recognition, and data mining communities \cite{gao2017sparse,towne2016measuring}.
A variety of similarity metrics, e.g., Cosine, Jaccard coefficient, Euclidean distance, Gaussian function, are often used in practice for convenience. However, they are often data-dependent and sensitive to noise \cite{huang2015new}. Consequently, different metrics lead to a big difference in the final results. In addition, several other similarity measure strategies are popular in dimension reduction techniques. For example, in the widely used locally linear embedding (LLE) \cite{roweis2000nonlinear}, isomeric feature mapping (ISOMAP) \cite{tenenbaum2000global}, and locality preserving projection (LPP) \cite{niyogi2004locality} methods, one has to construct an adjacency graph of neighbors. Then, k-nearest-neighborhood (knn) and $\epsilon$-nearest-neighborhood graph construction methods are often utilized. These approaches also have some inherent drawbacks, including 1) how to determine neighbor number $k$ or radius $\epsilon$; 2) how to choose an appropriate similarity metric to define neighborhood; 3) how to counteract the adverse effect of noise and outliers; 4) how to tackle data with structures at different scales of size and density. Unfortunately, all these factors heavily influence the subsequent tasks \cite{kang2018unified}. 

Recently, automatically learning of similarity information from data has drawn significant attention. In general, it can be classified into two categories. The first one is adaptive neighbors approach. It learns similarity information by assigning a probability for each data point as the neighborhood of another data point \cite{nie2014clustering}. It has been shown to be an effective way to capture the local manifold structure. 

The other one is self-expression approach. The basic idea is to represent every data point by a linear combination of other data points. In contrast, LLE reconstructs the original data by expressing each data point as a linear combination of its $k$ nearest neighbors only. Through minimizing this reconstruction error, we can obtain a coefficient matrix, which is also named similarity matrix. It has been widely applied in various representation learning tasks, including sparse subspace clustering \cite{elhamifar2013sparse,peng2016deep}, low-rank representation \cite{liu2013robust}, multi-view learning \cite{taoensemble2017}, semi-supervised learning \cite{zhuang2017label},  nonnegative matrix factorization(NMF) \cite{zhang2017adaptive}. 

However, this approach just tries to reconstruct the original data and has no explicit mechanism to preserve manifold structure information about the data. In many applications, the data can display structures beyond simply being low-rank or sparse. It is well-accepted that it is essential to take into account structure information when we perform high-dimensional data analysis. For instance, LLE preserves the local structure information. 

In view of this issue with the current approaches, we propose to learn the similarity information through reconstructing the original data kernel matrix, which is supposed to preserve overall relations. By doing so, we expect to obtain more accurate and complete data similarity. Considering clustering as a specific application of our proposed similarity learning method, we demonstrate that our framework provides impressive performance on several benchmark data sets. 
 In summary, the main contributions of this paper are threefold:
\begin{itemize}
\item{Compared to other approaches, the use of the kernel-based distances allows to work on preserving the sets of overall relations rather than individual pairwise similarities. }
\item{Similarity preserving provides a fundamental building block to embed high-dimensional data into low-dimensional latent space. It is general enough to be applied to a variety of learning problems.  }
\item{We evaluate the proposed approach in the clustering task. It shows that our algorithm enjoys superior performance compared to many state-of-the-art methods. }
\end{itemize}
\textbf{Notations.} Given a data set $\{x_1,x_2,\cdots,x_n\}$, we denote $X\in\mathcal{R}^{m \times n}$ with $m$ features and $n$ instances. Then the $(i,j)$-th element of matrix $X$ are denoted by $x_{ij}$. The $\ell_2$-norm of a vector \textbf{$x$} is represented by \textbf{$\|x\|=\sqrt{x^T\cdot x}$}, where $T$ denotes transpose. The $\ell_1$-norm of $X$ is defined as $\|X\|_1=\sum_{ij}|x_{ij}|$. The squared Frobenius norm is represented as $\|X\|_F^2=\sum_{ij}x_{ij}^2$. The nuclear norm of $X$ is $\|X\|_*=\sum\sigma_i$, where $\sigma_i$ is the $i$-th singular value of $X$. $I$ is the identity matrix with a proper size. $\vec{1}$ represents a column vector whose every element is one. $Z\geq 0$ means all the elements of $Z$ are nonnegative. Inner product $<x_i,x_j>=x_i^T\cdot x_j$.

\section{Related Work}
In this section, we provide a brief review of existing automatic similarity learning techniques.
\subsection{Adaptive Neighbors Approach}
 In a similar spirit of LPP, for each data point $x_i$, all the data points $\{x_j\}_{j=1}^n$ can be regarded as the neighborhood of $x_i$ with probability $z_{ij}$. To some extent, $z_{ij}$ represents the similarity between $x_i$ and $x_j$ \cite{nie2014clustering}. The smaller distance $\|x_i-x_j\|^2$ is, the greater the probability $z_{ij}$ is. Rather than prespecifying $Z$ with the deterministic neighborhood relation as LPP does, one can adaptively learn $Z$ from the data set by solving an optimization problem:  
\begin{equation}
\min_{z_i} \sum_{j=1}^n (\| x_i-x_j\|^2z_{ij}+\alpha z_{ij}^2)\hspace{.1cm} s.t.\hspace{.1cm} z_i^T\vec{1}=1, \hspace{.1cm} 0\leq z_{ij}\leq 1,
\label{local}
\end{equation}
where $\alpha$ is the regularization parameter. Recently, a variety of algorithms have been developed by using Eq. (\ref{local}) to learn a similarity matrix. Some applications are clustering \cite{nie2014clustering}, NMF \cite{huang2018robust}, and feature selection \cite{du2015unsupervised}. This approach can effectively capture the local structure information.

\subsection{Self-expression Approach}
The so-called self-expression is to approximate each data point as a linear combination of other data points, i.e., $x_i=\sum_j x_{j}z_{ij}$. The rationale here is that if $x_i$ and $x_j$ are similar, weight $z_{ij}$ should be big. Therefore, $Z$ also behaves like the similarity matrix. This shares the similar spirit as LLE, except that we do not predetermine the neighborhood. Its corresponding learning problem is:
\begin{equation}
\min_{Z}\frac{1}{2} \|X-XZ\|_F^2+\alpha \rho(Z)\quad
s.t.\quad Z\geq 0,
\label{global}
\end{equation} 
where $\rho(Z)$ is a regularizer of $Z$. Two commonly used assumptions about $Z$ are low-rank and sparse. Hence, in many domains, we also call $Z$ as the low-dimensional representation of $X$. Through this procedure, the individual pairwise similarity information hidden in the data is explored \cite{nie2014clustering} and the most informative ``neighbors'' for each data point are automatically chosen.  

Moreover, this learned $Z$ can not only reveal low-dimensional structure of data, but also be robust to data scale \cite{huang2015new}. Therefore, this approach has drawn significant attention and achieved impressive performance in a number of applications, including face recognition \cite{zhang2011sparse}, subspace clustering \cite{liu2013robust,elhamifar2013sparse},  semi-supervised learning \cite{zhuang2017label}. In many real-world applications, data often present complex structures. Nevertheless, the first term in Eq. (\ref{global}) simply minimizes the reconstruction error. Some important manifold structure information, such as overall relations, could be lost during this process. Preserving relation information has been shown to be important for feature selection \cite{zhao2013similarity}. In \cite{zhao2013similarity}, new feature vector $f$ is obtained by maximizing $f^T\hat{K}f$, where $\hat{K}$ is the refined similarity matrix derived from original kernel matrix $K$ with element $K(x,y)=\phi(x)^T\phi(y)$. In this paper, we propose a novel model to preserve the overall relations of the original data and simultaneously learn the similarity matrix. 

\section{Proposed Methodology} 

Since our goal is to obtain similarity information, it is very necessary to retain the overall relations among the data samples when we build a new representation. However, Eq. (\ref{global}) just tries to reconstruct the original data and does not take overall relations information into account. Our objective is finding a new representation which preserves overall relations as much as possible.

Given a data matrix $X$, one of the most commonly used relation measures is the inner product. Specifically, we try to minimize the inconsistency between two inner products: one for the raw data and another for reconstructed data $XZ$. To make our model more general, we build it in a transformed space,i.e., $X$ is mapped by $\phi$ \cite{xu2009extended}. We have
\begin{equation}
\label{pointsim}
\min_Z \|\phi(X)^T\cdot\phi(X)-(\phi(X)Z)^T\cdot(\phi(X)Z)\|_F^2
\end{equation}
 (\ref{pointsim}) can be simplified as 
\begin{equation}
\min_Z \|K-Z^T KZ\|_F^2.
\label{preserve}
\end{equation}

With certain assumption about the structure of $Z$, our proposed  \textbf{S}imilarity \textbf{L}earning via \textbf{K}ernel preserving \textbf{E}mbedding (SLKE) framework can be formulated as
\begin{equation}
\min_Z  \frac{1}{2}\|K-Z^TKZ\|_F^2+\gamma \rho(Z)\quad
s.t. \quad Z\ge 0,
\label{newobj}
\end{equation}
where $\gamma>0$ is a tradeoff parameter and $\rho$ is a regularizer on $Z$. If we use the nuclear norm $\|\cdot\|_*$ to replace $\rho(\cdot)$, we have a low-rank representation. If the $\ell_1$-norm is adopted, we obtain a sparse representation. 
 It is worth pointing out that Eq. (\ref{newobj}) enjoys several nice properties: \\
1) The use of kernel-based distance preserves the sets of overall relations, which will benefit the subsequent tasks;\\ 
2) This learned low-dimensional representation or similarity matrix $Z$ is general enough to be utilized to solve a variety of different tasks, where similarity information is needed; \\
3) The learned representation is particularly suitable to problems that are sensitive to data similarity, such as clustering \cite{kang2018low}, classification \cite{wright2009robust}, recommender systems \cite{kang2017kernel}; \\
4) Its input is the kernel matrix. This is also desirable, as not all types of data can be represented in numerical feature vectors form \cite{xu2010simple}. For instance, we need to group proteins in bioinformatics based on their structures and to divide users in social media based on their friendship relations.\\
In the following section, we will show a simple strategy to solve problem (\ref{newobj}).

\section{Optimization}
It is easy to see that Eq. (\ref{newobj}) is a fourth-order function of $Z$. Directly solving it is not so straightforward. To circumvent this problem, we first convert it to the following equivalent problem by introducing two more auxiliary variables
\begin{equation}
\begin{split}
\min_Z  \frac{1}{2}\|K-J^TKW\|_F^2+\gamma \rho(Z)\\
s.t. \quad Z\ge 0,\hspace{.1cm}Z=J,\hspace{.1cm} Z=W.
\label{obj}
\end{split}
\end{equation}
Now we resort to the alternating direction method of  multipliers (ADMM) method to solve (\ref{obj}). The corresponding augmented Lagrangian function is：
\begin{equation}
\begin{split}
\mathcal{L}(Z,J,W,Y_1,Y_2)=\frac{1}{2}\|K-J^TKW\|_F^2+\gamma \rho(Z)+\\
\frac{\mu}{2}\left(\|Z-J+\frac{Y_1}{\mu}\|_F^2+\|Z-W+\frac{Y_2}{\mu}\|_F^2\right),
\label{lag}
\end{split}
\end{equation}
where $\mu>0$ is a penalty parameter and $Y_1$, $Y_2$ are the lagrangian multipliers. The variables $Z$, $W$, and $J$ can be updated alternatingly, one at each step, while keeping the other two fixed. 

To solve $J$, we observe that the objective function (\ref{lag}) is a strongly convex quadratic function in $J$ which can be solved by setting its first derivative to zero, we have:
\begin{equation}
J=(\mu I+KWW^TK^T)^{-1}(\mu Z+Y_1+KWK^T),
\label{updateJ}
\end{equation} 
where $I\in \mathcal{R}^{n\times n}$ is the identity matrix.

Similarly, 
\begin{equation}
W=(\mu I+K^TJJ^TK)^{-1}(\mu Z+Y_2+K^TJK).
\label{updateW}
\end{equation} 

For $Z$, we have the following subproblem:
\begin{equation}
\min_Z  \gamma\rho(Z)+\mu \left\|Z-\frac{J+W-\frac{Y_1+Y_2}{\mu}}{2}\right\|_F^2.
\end{equation}
Depending on different regularization strategies, we have different closed-form solutions for $Z$. Define $H=\frac{J+W-\frac{Y_1+Y_2}{\mu}}{2}$, we can write its singular value decomposition (SVD) as $Udiag(\sigma)V^T$. Then, for low-rank representation, i.e., $\rho(Z)=\|Z\|_*$, we have,
 \begin{equation}
 Z=U diag(max\{\sigma-\frac{\gamma}{2\mu},0\}) V^T.
\label{lowrank}
\end{equation}

For sparse representation, i.e., $\rho(Z)=\|Z\|_1$, we can update $Z$ element by element as,
\begin{equation}
 Z_{ij}=max\{|H_{ij}|-\frac{\gamma}{2\mu},0\}\cdot sign(H_{ij}).
\label{sparse}
\end{equation}
For clarity, the complete procedures to solve the problem (\ref{newobj}) are outlined in Algorithm 1.

\begin{algorithm}

\caption{The algorithm of SLKE}
\label{alg2}
 {\bfseries Input:} Kernel matrix $K$, parameters  $\gamma>0$,  $\mu>0$.\\
{\bfseries Initialize:} Random matrix $W$ and $Z$, $Y_1=Y_2=0$.\\
 {\bfseries REPEAT}
\begin{algorithmic}[1]
\STATE Calculate $J$ by (\ref{updateJ}).
 \STATE Update $W$ according to (\ref{updateW}).
\STATE Calculate $Z$ using (\ref{lowrank}) or (\ref{sparse}).
\STATE Update Lagrange multipliers $Y_1$ and $Y_2$ as
\begin{eqnarray*}
Y_1=Y_1+\mu(Z-J),\\
Y_2=Y_2+\mu(Z-W).
\end{eqnarray*}
\end{algorithmic}
\textbf{ UNTIL} {stopping criterion is met.}
\end{algorithm}
\subsection{Complexity Analysis}

With our optimization strategy, the complexity for $J$ is $\mathcal{O}(n^3)$. Updating $W$ has the same complexity as $J$. Both $J$ and $W$ involve matrix inverse. Fortunately, we can avoid it by resorting to some approximation techniques when we face large-scale data sets. Depending on the choice of regularizer, we have different complexity for $Z$. For low-rank representation, it requires an SVD for every iteration and its complexity is $\mathcal{O}(n^3)$. Since we seek a low-rank matrix and so only need a few principle singular values. Package like PROPACK can compute a rank $k$ SVD with complexity $\mathcal{O}(n^2k)$ \cite{larsen2004propack}. To obtain a sparse solution of $Z$, we need $\mathcal{O}(n^2)$ complexity. The updating of $Y_1$ and $Y_2$ cost $\mathcal{O}(n^2)$.
\section{Experiments}
To assess the effectiveness of our proposed method, we apply the learned similarity matrix to do clustering. 
\subsection{Data Sets}
\begin{table}[!htbp]
\centering
\caption{Description of the data sets}
\label{data}

\begin{tabular}{|l|c|c|c|}
\hline
&\textrm{\# instances}&\textrm{\# features}&\textrm{\# classes}\\\hline
\textrm{YALE}&165&1024&15\\\hline
\textrm{JAFFE}&213&676&10\\\hline
\textrm{ORL}&400&1024&40\\\hline
\textrm{COIL20}&1440&1024&20\\\hline
\textrm{BA}&1404&320&36\\\hline
\textrm{TR11}&414&6429&9\\\hline
\textrm{TR41}&878&7454&10\\\hline
\textrm{TR45}&690&8261&10\\\hline
\textrm{TDT2}&9394&36771&30\\\hline
\end{tabular}
\end{table}

We conduct our experiments with nine benchmark data sets, which are widely used in clustering experiments. We show the statistics of these data sets in Table \ref{data}. In summary, the number of data samples varies from 165 to 9,394 and feature number ranges from 320 to 36,771. The first five data sets are images, while the last four are text data. 

Specifically, the five image data sets contain three face databases  (ORL, YALE, and JAFFE), a toy image database COIL20, and a binary alpha digits data set BA. For example, COIL20 consists of 20 objects and each object  was taken from different angles.
BA data set contains images of digits of ``0" through ``9" and letters of capital ``A" through ``Z". YALE, ORL, and JAFEE consist of images of the person. Each image represents different facial expressions or configurations due to times, illumination conditions, and glasses/no glasses.
\subsection{Data Preparation}
Since the input for our proposed method is kernel matrix, we design 12 kernels in total to fully examine its performance. They are: seven Gaussian kernels of the form $K(x,y)=exp(-\|x-y\|_2^2/(td_{max}^2))$ with $t\in \{0.01, 0.05, 0.1, 1, 10, 50, 100\}$, where $d_{max}$ denotes the maximal distance between data points; a linear kernel $K(x,y)=x^T y$; four polynomial kernels $K(x,y)=(a+x^T y)^b$ of the form with $a\in\{0,1\}$ and $b\in\{2,4\}$. Besides, all kernels are rescaled to $[0,1]$ by dividing each element by the largest element in its corresponding kernel matrix. These kernels are commonly used types in the literature, so we can well investigate the performance of our method.
\subsection{Comparison Methods}
To fully examine the effectiveness of the proposed framework on clustering, we choose a good set of methods to compare. In general, they can be classified into two categories: similarity-based and kernel-based clustering methods.\\
\begin{itemize}
\item{\textbf{Spectral Clustering (SC) }\cite{ng2002spectral}: SC is a widely used clustering method. It enjoys the advantage of exploring the intrinsic data structures. Its input is the graph Laplacian, which is constructed from the similarity matrix. Here, we directly treat kernel matrix as the similarity matrix for spectral clustering. For our proposed SLKE method, we employ learned $Z$ to do spectral clustering. Thus, SC serves as a baseline method.}
\item{\textbf{Robust Kernel K-means (RKKM)\footnote{\label{note1}https://github.com/csliangdu/RMKKM}} \cite{du2015robust}: Based on classical k-means clustering algorithm, RKKM has been developed to deal with nonlinear structures, noise, and outliers in the data. RKKM demonstrates superior performance on a number of real-world data sets. }
\item{\textbf{Simplex Sparse Representation (SSR)} \cite{huang2015new}: SSR method has been proposed recently. It is based on adaptive neighbors idea. Another appealing property of this method is that its model parameter can be calculated by assuming a maximum number of neighbors. Therefore, we don't need to tune the parameter any more. In addition, it outperforms many other state-of-the-art techniques. }
\item{\textbf{Low-Rank Representation (LRR)} \cite{liu2013robust}: Based on self-expression, subspace clustering with low-rank regularizer achieves great success on a number of applications, such as face clustering, motion segmentation.}
\item{\textbf{Sparse Subspace Clustering (SSC)} \cite{elhamifar2013sparse}: Similar to LRR, SSC assumes a sparse solution of $Z$. Both LRR and SSC learn similarity matrix by reconstructing the original data. In this aspect, SC, LRR, and SSC are baseline methods w.r.t. our proposed algorithm. }
\item \textbf{Clustering with Adaptive Neighbor (CAN)} \cite{nie2014clustering}. Based on the idea of adaptive neighbors, i.e., Eq.(\ref{local}), CAN learns a local graph from raw data for clustering task. 
\item{\textbf{Twin Learning for Similarity and Clustering (TLSC)} \cite{kang2017twin}: Recently, TLSC has been proposed and has shown promising results on real-world data sets. TLSC does not only learn similarity matrix via self-expression in kernel space, but also have optimal similarity graph guarantee. Besides, it has good theoretical properties, i.e., it is equivalent to kernel k-means and k-means under certain conditions. }
\item{\textbf{SLKE}: Our proposed similarity learning method with overall relations preserving capability. After obtaining similarity matrix $Z$, we use spectral clustering to conduct clustering experiments. We test both low-rank and sparse regularizer. We denote them as SLKE-R and SLKE-S, respectively\footnote{https://github.com/sckangz/SLKE}.  }
\end{itemize}
\subsection{Evaluation Metrics}
To quantitatively and effectively assess the clustering performance, we utilize the two widely used metrics \cite{peng2018integrate}, accuracy (Acc) and normalized mutual information (NMI).

Acc discovers the one-to-one relationship between clusters and classes. Let $l_i$ and $\hat{l}_i$ be the clustering result and the ground truth cluster label of $x_i$, respectively. Then the Acc is defined as
\[
Acc=\frac{\sum_{i=1}^n \delta(\hat{l}_i, map(l_i))}{n},
\]
where $n$ is the sample size, Kronecker delta function $\delta(x,y)$ equals one if and only if $x=y$ and zero otherwise, and map($\cdot$) is the best permutation mapping function that maps each cluster index to a true class label  based on Kuhn-Munkres algorithm. 

 Given two sets of clusters $L$ and $\hat{L}$, NMI is defined as
\[
\textrm{NMI}(L,\hat{L})=\frac{\sum\limits_{l\in L,\hat{l}\in\hat{L}} p(l,\hat{l})\textrm{log}(\frac{p(l,\hat{l})}{p(l)p(\hat{l})})}{\textrm{max}(H(L),H(\hat{L}))},
\]  
where $p(l)$ and $p(\hat{l})$ represent the marginal probability distribution functions of $L$ and $\hat{L}$, respectively. $p(l,\hat{l})$ is the joint probability function of $L$ and $\hat{L}$. $H(\cdot)$ is the entropy function. The greater NMI means the better clustering performance.

\subsection{Results}
\captionsetup{position=top}
\begin{table*}[!ht]
\centering
\subfloat[Accuracy(\%)\label{acc}]{
\begin{tabular}{|l  |c | c| c| c|c|c |c| c|  c|}
	\hline
    \small{Data}  & \small{SC} &\small{RKKM} & \small{SSC} & \small{LRR}&\small{SSR}&\small{CAN} & \small{TLSC}  &\small{SLKE-S}&\small{SLKE-R}  \\
     \hline
        \multirow{1}{*}{\small{YALE}}   &49.42(40.52)&48.09(39.71)&38.18&61.21&54.55&58.79&55.85(45.35)&61.82(38.89)&\textbf{66.24}(51.28)\\
	
		\hline
		\multirow{1}{*}{\small{JAFFE}}  & 74.88(54.03)&75.61(67.89)&99.53&99.53&87.32&98.12&99.83(86.64)&96.71(70.77)&\textbf{99.85}(90.89)\\
		
		\hline
        \multirow{1}{*}{\small{ORL}}  & 58.96(46.65)&54.96(46.88)&36.25&76.50&69.00&61.50&62.35(50.50)&\textbf{77.00}(45.33)&74.75(59.00)\\

		\hline
       \multirow{1}{*}{\small{COIL20}}  & 67.60(43.65)&61.64(51.89)&73.54&68.40&76.32&\textbf{84.58}&72.71(38.03)&75.42(56.83)&84.03(65.65)\\
		
		\hline
        \multirow{1}{*}{\small{BA}}& 31.07(26.25)&42.17(34.35)&24.22&45.37&23.97&36.82&47.72(39.50)&\textbf{50.74}(36.35)&44.37(35.79)\\
    
      \hline
       \multirow{1}{*}{\small{TR11}}  & 50.98(43.32)&53.03(45.04)&32.61&73.67&41.06&38.89&71.26(54.79)&69.32(46.87)&\textbf{74.64}(55.07)\\
		
		\hline
		\multirow{1}{*}{\small{TR41}}  & 63.52(44.80)&56.76(46.80)&28.02&70.62&63.78&62.87&65.60(43.18)&71.19(47.91)&\textbf{74.37}(53.51)\\
		
		\hline
		\multirow{1}{*}{\small{TR45}} & 57.39(45.96)&58.13(45.69)&24.35&78.84&71.45&48.41&74.02(53.38)&78.55(50.59)&\textbf{79.89}(58.37)\\
		
		\hline
		\multirow{1}{*}{\small{TDT2}}&52.63(45.26)&48.35(36.67)&23.45&52.03&20.86&19.74 &55.74(44.82)&59.61(25.40)&\textbf{74.92}(33.67)\\
\hline
\end{tabular}
}

\subfloat[NMI(\%)\label{NMI}]{

\begin{tabular}{|l  |c | c| c| c|c|c |c| c| c|}
	\hline
    \small{Data}  & \small{SC} &\small{RKKM} & \small{SSC} & \small{LRR}&\small{SSR} &\small{CAN}& \small{TLSC}  & \small{SLKE-S}&\small{SLKE-R}  \\
       \hline
        \multirow{1}{*}{\small{YALE}} & 52.92(44.79)&52.29(42.87)&45.56&62.98&57.26&57.67&56.50(45.07)&59.47(40.38)&\textbf{64.29}(52.87)\\
		
		\hline
		\multirow{1}{*}{\small{JAFFE}} & 
82.08(59.35)&83.47(74.01)&99.17&99.16&92.93&97.31&99.35(84.67)&94.80(60.83)&\textbf{99.49}(81.56)\\
		
	\hline
        \multirow{1}{*}{\small{ORL}} & 75.16(66.74)&74.23(63.91)&60.24&85.69&84.23&76.59&78.96(63.55)&\textbf{86.35}(58.84)&85.15(75.34)\\

		\hline
       \multirow{1}{*}{\small{COIL20}}  &80.98(54.34)&74.63(63.70)&80.69&77.87&86.89&\textbf{91.55}&82.20(73.26)&80.61(65.40)&91.25(73.53)\\
		
	\hline
        \multirow{1}{*}{\small{BA}}  &50.76(40.09)&57.82(46.91)&37.41&57.97&30.29&49.32&63.04(52.17)&\textbf{63.58}(55.06)&56.78(50.11)\\
      
	\hline
        \multirow{1}{*}{\small{TR11}}  & 43.11(31.39)&49.69(33.48)&02.14&65.61&27.60&19.17&58.60(37.58)&67.63(30.56)&\textbf{70.93}(45.39)\\
		\hline
 \multirow{1}{*}{\small{TR41}}  &61.33(36.60)&60.77(40.86)&01.16&67.50&59.56&51.13&65.50(43.18)&\textbf{70.89}(34.82)&68.50(47.45)\\
		
		\hline
		\multirow{1}{*}{\small{TR45}}& 48.03(33.22)&57.86(38.96)&01.61&77.01&67.82&49.31&74.24(44.36)&72.50(38.04)&\textbf{78.12}(50.37)\\
	\hline
\multirow{1}{*}{\small{TDT2}}&52.23(27.16)&54.46(42.19)&13.09&64.36&02.44&03.97&58.35(46.37)&58.55(15.43)&\textbf{68.21}(28.94)\\
\hline
\end{tabular}}
\caption{Clustering results obtained on benchmark data sets. The average performance of those 12 kernels is put in parenthesis. The best results among those kernels are highlighted in boldface. \label{clusterres}}
\end{table*}

We report the extensive experimental results in Table \ref{clusterres}. Except SSC, LRR, CAN, and SSR, we run other methods on each kernel matrix individually. As a result, we show both the best  performance among those 12 kernels and the average results over those 12 kernels for them. Based on this table, we can see that our proposed SLKE achieves the best performance in most cases. To be specific, we have the following observations:\\
 1) Compared to classical k-means based RKKM and spectral clustering techniques, our proposed method SLKE has a big advantage in terms of accuracy and NMI. With respect to the recently proposed SSR and TLSC methods, SKLE always obtains better results.\\
2) SLKE-R and SLKE-S often outperform LRR and SSC, respectively. The accuracy increased by 8.92\%, 8.76\% on average, respectively. That is to say, kernel-based distance approach indeed performs better than original data reconstruction technique. This verifies the importance of retaining relation information when we learn a low-dimensional representation, especially for sparse representation.  \\

3) With respect to adaptive neighbors approach CAN, we also obtain better performance on those datasets except COIL20. For COIL20, our results are quite close to CAN's. Therefore, compared to various similarity learning techniques, our method is very competitive.\\
4) Regarding low-rank and sparse representation, it is hard to conclude which one is better. It totally depends on the specific data.

\begin{figure}[!hbtp]
\centering
\subfloat[Original data]{\includegraphics[width=.49\textwidth]{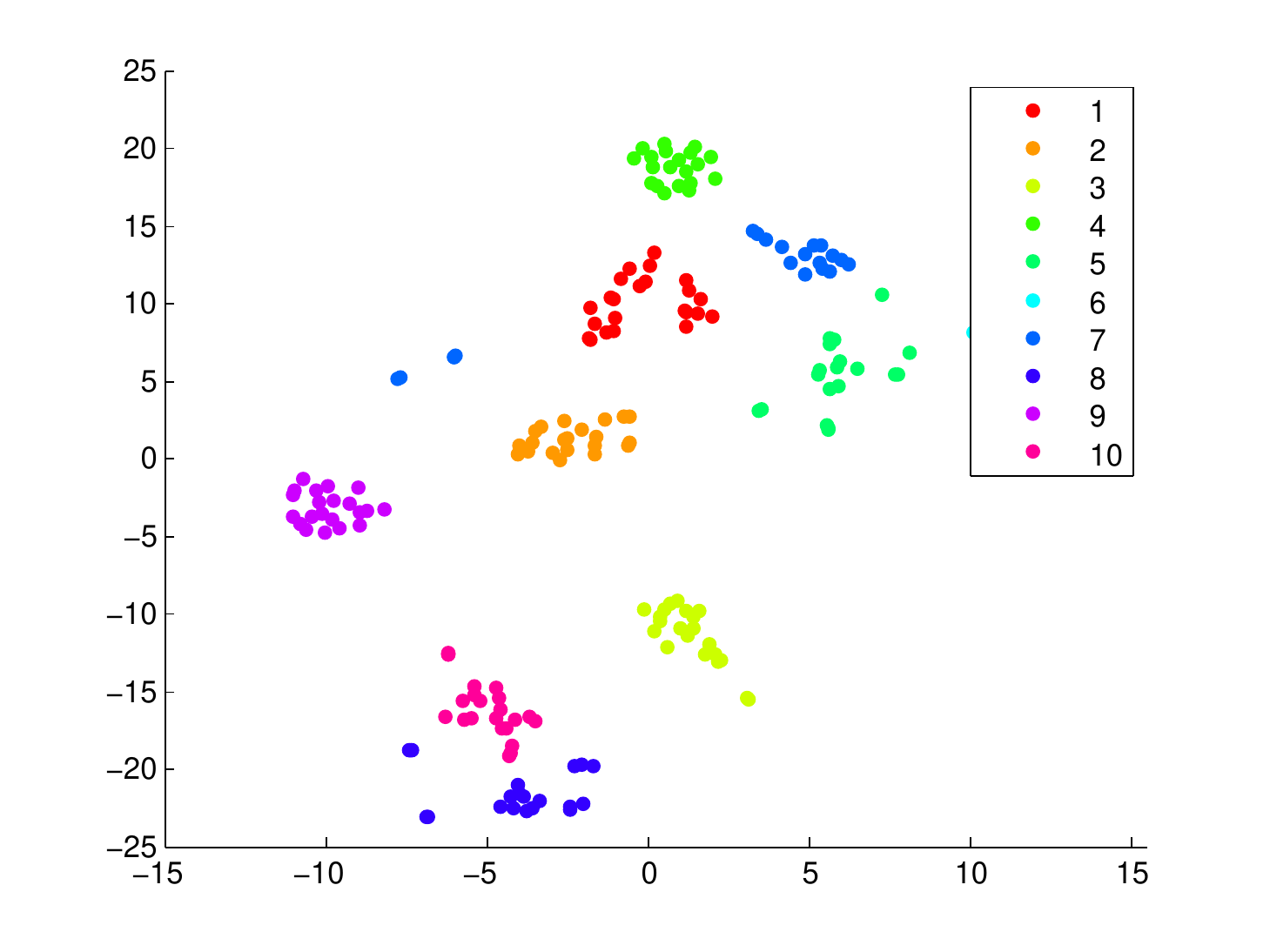}}\\
\subfloat[Reconstructed data]{\includegraphics[width=.49\textwidth]{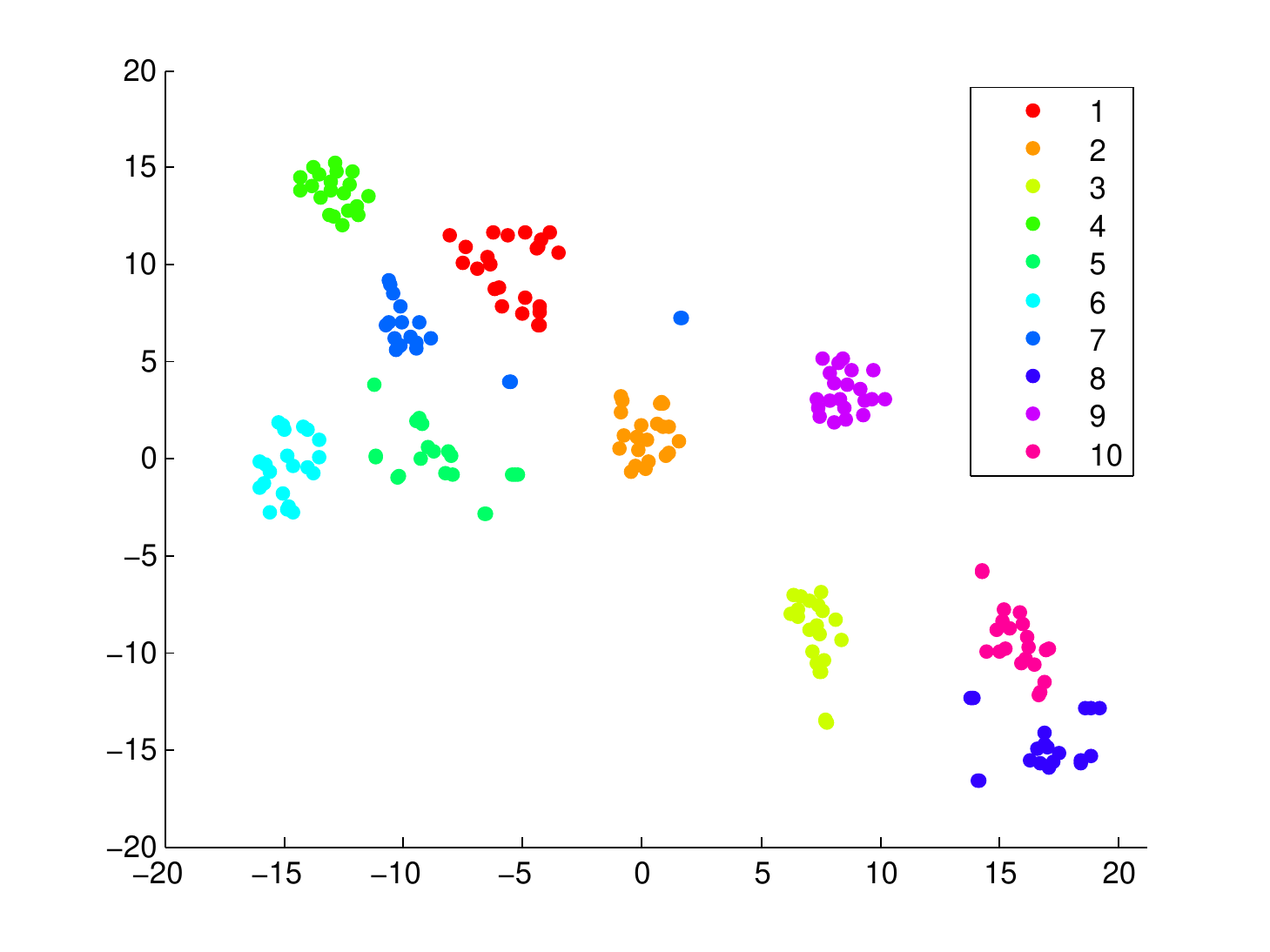}}
\caption{JAFFE data set visualized in two dimensions.\label{visual}}
\end{figure}

Furthermore, we run t-SNE \cite{maaten2008visualizing} algorithm on the JAFFE data $X$ and the reconstructed data $XZ$ from the best result of our SLKE-R. As shown by Figure \ref{visual}, we can see that our method can well preserve the cluster structure of the data. 
\captionsetup{position=bottom}

\captionsetup{position=bottom}
\begin{figure*}[!hbtp]
\centering
\subfloat[Acc\label{acc}]{\includegraphics[width=.48\textwidth]{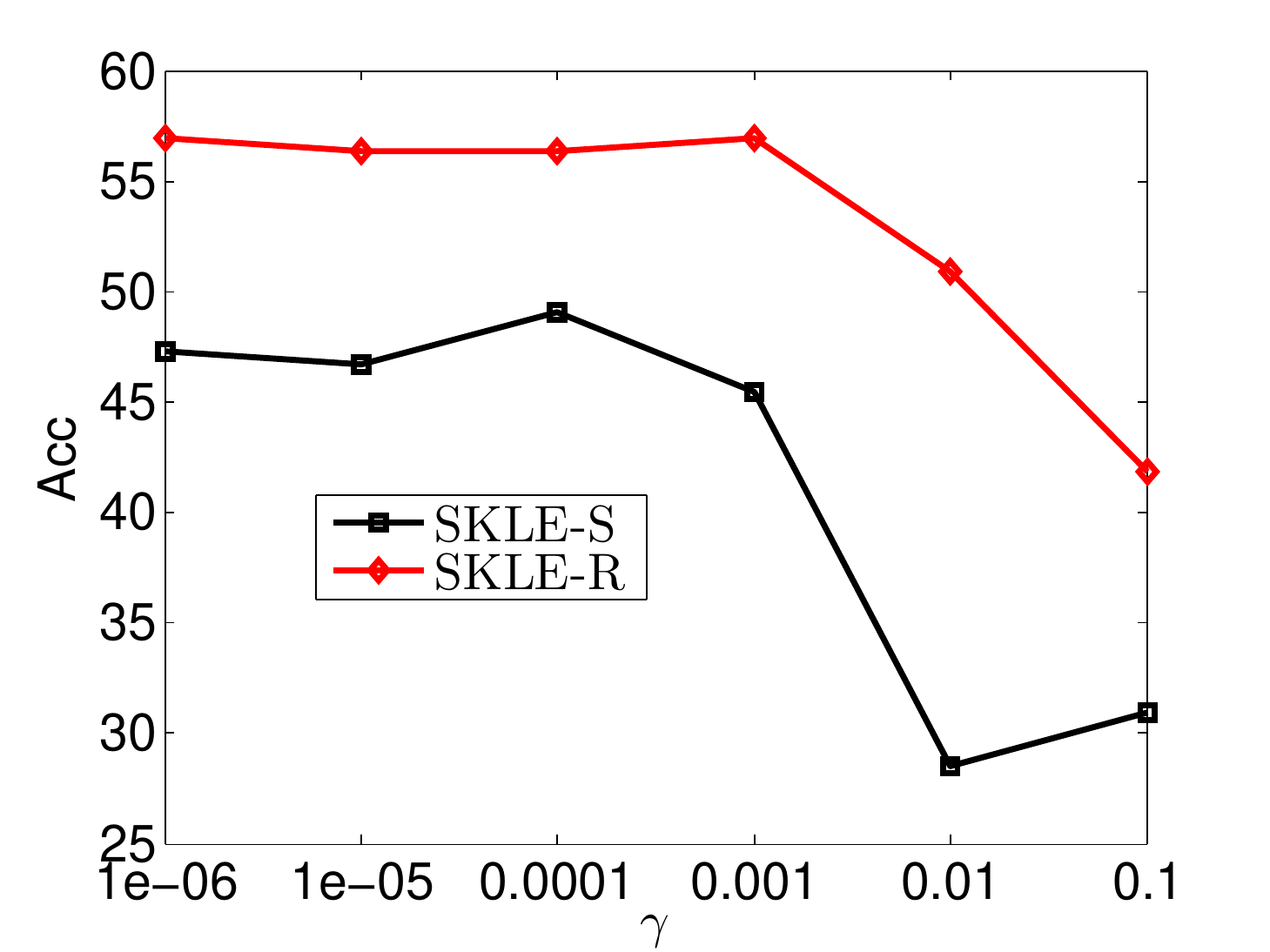}}
\subfloat[NMI\label{nmi}]{\includegraphics[width=.48\textwidth]{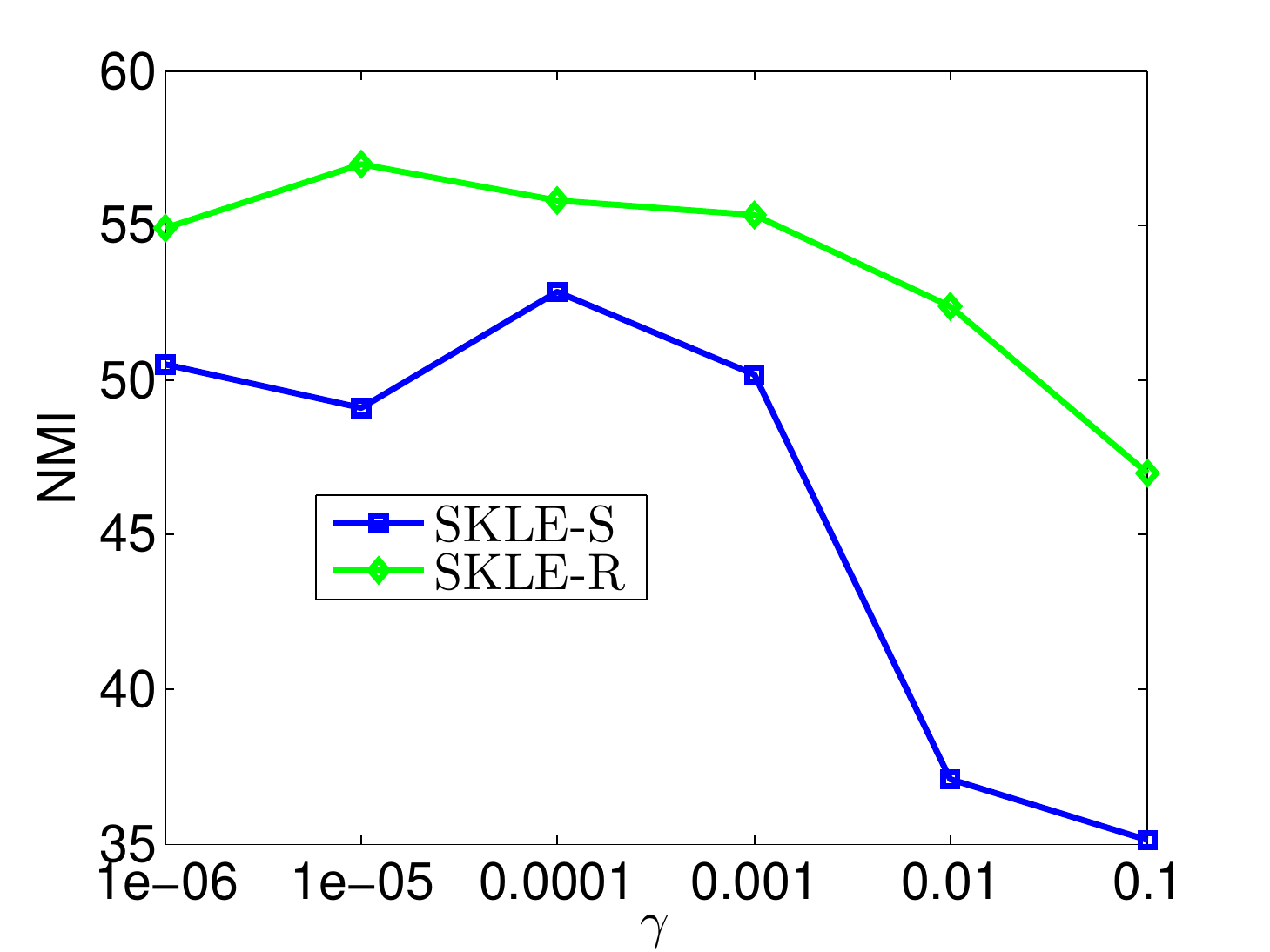}}
\caption{The effect of parameter $\gamma$ on the YALE data set.\label{yalepara}}
\end{figure*}
\begin{table*}[!htbp]
\centering
\caption{Wilcoxon Signed Rank Test on all Data sets.}
\label{Wilcoxon}

\begin{tabular}{|c|c|c|c|c|c|c|c|c|}
\hline
\textrm{Method}&\textrm{Metric}&\textrm{SC}&\textrm{RKKM}&\textrm{SSC}&\textrm{LRR}&\textrm{SSR}&\textrm{CAN}&\textrm{TLSC}\\\hline

\multirow{2}{4em}{\textrm{SLKE-S}}&Acc&.0039&.0039&.0117&.2500&.0078&.0391&.0391\\ 
\cline{2-9}
&NMI&.0078&.0039&.0195&.6523&.0391&.0547&.3008\\ \hline
\multirow{2}{4em}{\textrm{SLKE-R}}&Acc&.0039&.0039&.0039&.0977&.0039&.0039&.0117\\
\cline{2-9}
&NMI&.0039&.0078&.0039&.0742&.0039&.0078&.0391\\ \hline
\end{tabular}
\end{table*}
To see the significance of improvements, we further apply the Wilcoxon signed rank test \cite{peng2017supervised} to Table \ref{clusterres}. We show the $p$-values in Table \ref{Wilcoxon}. We note that the testing results are under 0.05 in most cases when comparing SLKE-S and SLKE-R to other methods. Therefore, SLKE-S and SLKE-R outperform SC, RKKM, SSC, and SSR with statistical significance.

\subsection{Parameter Analysis}
\begin{figure}[!hbtp]
\centering
\subfloat[Acc\label{purity}]{\includegraphics[width=.45\textwidth]{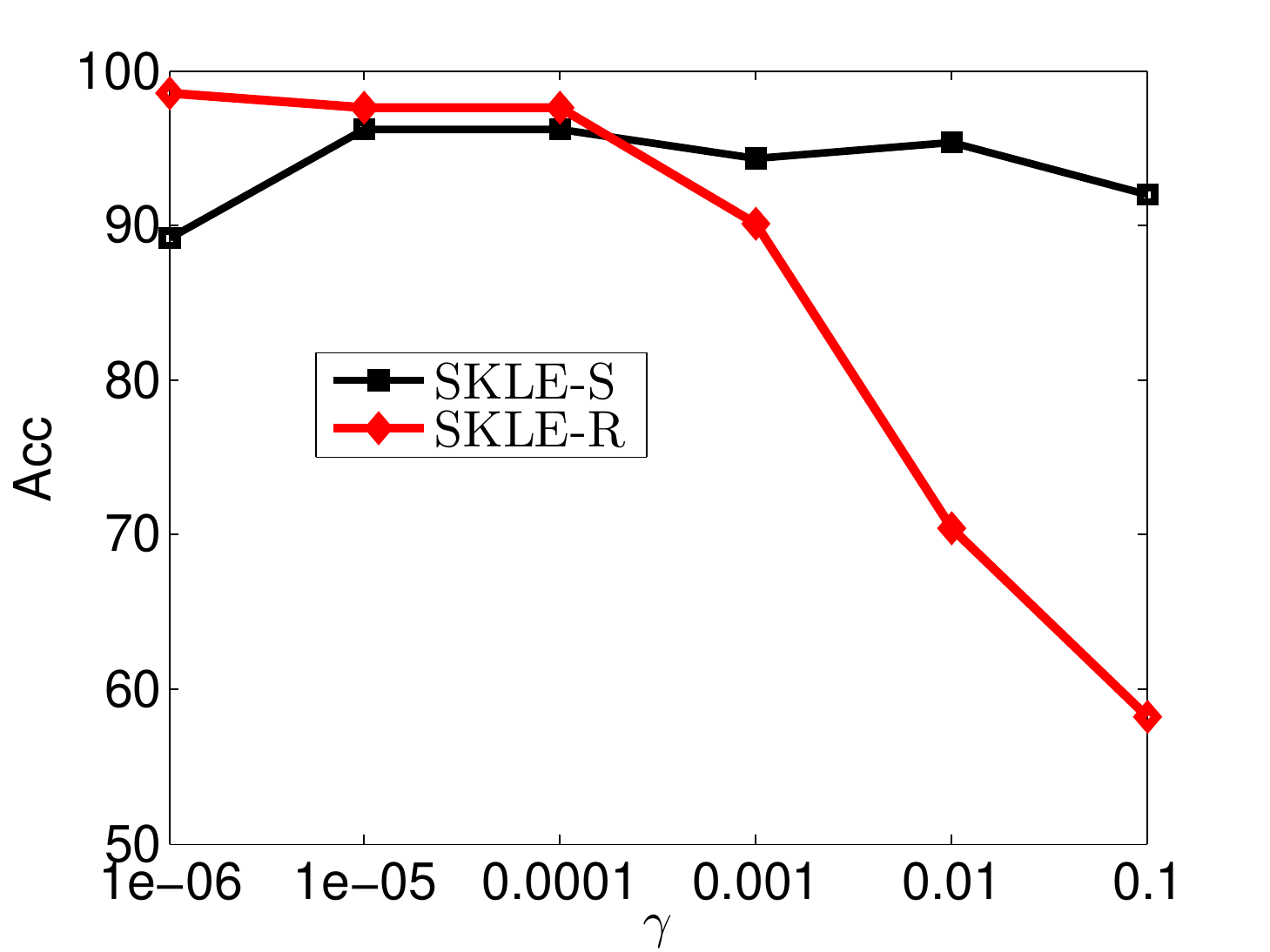}}\\
\subfloat[NMI\label{purity}]{\includegraphics[width=.45\textwidth]{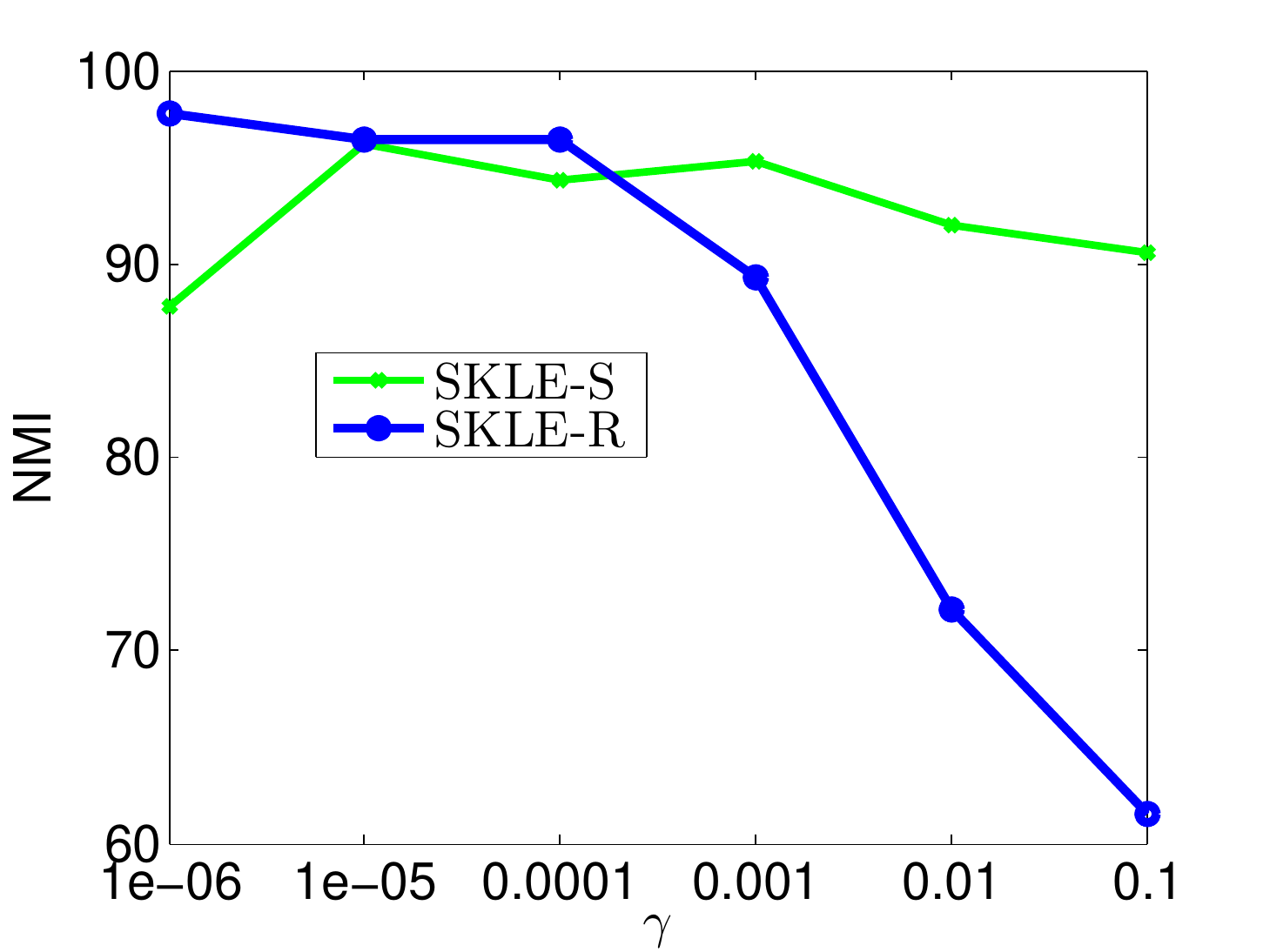}}
\caption{The effect of parameter $\gamma$ on the JAFFE data set.\label{jaffepara}}
\end{figure}

In this subsection, we investigate the influence of our model parameter $\gamma$ on the clustering results. Take Gaussian kernel with $t=100$ of YALE and JAFFE data sets as examples, we plot our algorithm's performance with $\gamma$ in the range $[10^{-6}, 10^{-5}, 10^{-4}, 10^{-3}, 10^{-2}, 10^{-1} ]$ in Figure \ref{yalepara} and \ref{jaffepara}, respectively. As we can see that our proposed methods work well for a wide range of $\gamma$, e.g., from $10^{-6}$ to $10^{-3}$. 

\section{Conclusion}
In this paper, we present a novel similarity learning framework relying on an embedding of kernel-based distance. Our model is flexible to obtain either low-rank or sparse representation of data. Comprehensive experimental results on real data sets well demonstrate the superiority of the proposed method on the clustering task. It has great potential to be applied in a number of applications beyond clustering.   
It has been shown that the performance of the proposed method is largely determined by the choice of kernel function. In the future, we plan to address this issue by developing a multiple kernel learning method, which is capable of automatically learning an appropriate kernel from a pool of input kernels. 

\section{ACKNOWLEDGMENT}
This paper was in part supported by Grants from the Natural
Science Foundation of China (Nos. 61806045 and 61572111),
a 985 Project of UESTC (No. A1098531023601041) and two
Fundamental Research Fund for the Central Universities of
China (Nos. A03017023701012 and ZYGX2017KYQD177).


\begin{thebibliography}{}

\bibitem[\protect\citeauthoryear{Chen \bgroup et al\mbox.\egroup
  }{2012}]{chen2012fgkm}
Chen, X.; Ye, Y.; Xu, X.; and Huang, J.~Z.
\newblock 2012.
\newblock A feature group weighting method for subspace clustering of
  high-dimensional data.
\newblock {\em Pattern Recognition} 45(1):434--446.

\bibitem[\protect\citeauthoryear{Chen \bgroup et al\mbox.\egroup
  }{2018}]{chen2017dnc}
Chen, X.; Hong, W.; Nie, F.; He, D.; Yang, M.; and Huang, J.~Z.
\newblock 2018.
\newblock Directly minimizing normalized cut for large scale data.
\newblock In {\em SIGKDD},  1206--1215.

\bibitem[\protect\citeauthoryear{Du and Shen}{2015}]{du2015unsupervised}
Du, L., and Shen, Y.-D.
\newblock 2015.
\newblock Unsupervised feature selection with adaptive structure learning.
\newblock In {\em SIGKDD},  209--218.
\newblock ACM.

\bibitem[\protect\citeauthoryear{Du \bgroup et al\mbox.\egroup
  }{2015}]{du2015robust}
Du, L.; Zhou, P.; Shi, L.; Wang, H.; Fan, M.; Wang, W.; and Shen, Y.-D.
\newblock 2015.
\newblock Robust multiple kernel k-means using ℓ 2; 1-norm.
\newblock In {\em IJCAI},  3476--3482.
\newblock AAAI Press.

\bibitem[\protect\citeauthoryear{Elhamifar and
  Vidal}{2013}]{elhamifar2013sparse}
Elhamifar, E., and Vidal, R.
\newblock 2013.
\newblock Sparse subspace clustering: Algorithm, theory, and applications.
\newblock {\em IEEE transactions on pattern analysis and machine intelligence}
  35(11):2765--2781.

\bibitem[\protect\citeauthoryear{Gao \bgroup et al\mbox.\egroup
  }{2017}]{gao2017sparse}
Gao, X.; Hoi, S.~C.; Zhang, Y.; Zhou, J.; Wan, J.; Chen, Z.; Li, J.; and Zhu,
  J.
\newblock 2017.
\newblock Sparse online learning of image similarity.
\newblock {\em ACM Transactions on Intelligent Systems and Technology (TIST)}
  8(5):64.

\bibitem[\protect\citeauthoryear{Hirzer \bgroup et al\mbox.\egroup
  }{2012}]{hirzer2012relaxed}
Hirzer, M.; Roth, P.~M.; K{\"o}stinger, M.; and Bischof, H.
\newblock 2012.
\newblock Relaxed pairwise learned metric for person re-identification.
\newblock In {\em ECCV},  780--793.
\newblock Springer.

\bibitem[\protect\citeauthoryear{Hoi, Liu, and Chang}{2008}]{hoi2008semi}
Hoi, S.~C.; Liu, W.; and Chang, S.-F.
\newblock 2008.
\newblock Semi-supervised distance metric learning for collaborative image
  retrieval.
\newblock In {\em CVPR},  1--7.
\newblock IEEE.

\bibitem[\protect\citeauthoryear{Huang \bgroup et al\mbox.\egroup
  }{2018}]{huang2018robust}
Huang, S.; Wang, H.; Li, T.; Li, T.; and Xu, Z.
\newblock 2018.
\newblock Robust graph regularized nonnegative matrix factorization for
  clustering.
\newblock {\em Data Mining and Knowledge Discovery} 32(2):483--503.

\bibitem[\protect\citeauthoryear{Huang, Nie, and Huang}{2015}]{huang2015new}
Huang, J.; Nie, F.; and Huang, H.
\newblock 2015.
\newblock A new simplex sparse learning model to measure data similarity for
  clustering.
\newblock In {\em IJCAI},  3569--3575.

\bibitem[\protect\citeauthoryear{Kang \bgroup et al\mbox.\egroup
  }{2018a}]{kang2018self}
Kang, Z.; Lu, X.; Yi, J.; and Xu, Z.
\newblock 2018a.
\newblock Self-weighted multiple kernel learning for graph-based clustering and
  semi-supervised classification.
\newblock In {\em IJCAI},  2312--2318.

\bibitem[\protect\citeauthoryear{Kang \bgroup et al\mbox.\egroup
  }{2018b}]{kang2018unified}
Kang, Z.; Peng, C.; Cheng, Q.; and Xu, Z.
\newblock 2018b.
\newblock Unified spectral clustering with optimal graph.
\newblock In {\em Proceedings of the Thirty-Second AAAI Conference on
  Artificial Intelligence (AAAI-18). AAAI Press}.

\bibitem[\protect\citeauthoryear{Kang \bgroup et al\mbox.\egroup
  }{2018c}]{kang2018low}
Kang, Z.; Wen, L.; Chen, W.; and Xu, Z.
\newblock 2018c.
\newblock Low-rank kernel learning for graph-based clustering.
\newblock {\em Knowledge-Based Systems}.

\bibitem[\protect\citeauthoryear{Kang, Peng, and Cheng}{2017a}]{kang2017kernel}
Kang, Z.; Peng, C.; and Cheng, Q.
\newblock 2017a.
\newblock Kernel-driven similarity learning.
\newblock {\em Neurocomputing} 267:210--219.

\bibitem[\protect\citeauthoryear{Kang, Peng, and Cheng}{2017b}]{kang2017twin}
Kang, Z.; Peng, C.; and Cheng, Q.
\newblock 2017b.
\newblock Twin learning for similarity and clustering: A unified kernel
  approach.
\newblock In {\em Proceedings of the Thirty-First AAAI Conference on Artificial
  Intelligence (AAAI-17). AAAI Press}.

\bibitem[\protect\citeauthoryear{Larsen}{2004}]{larsen2004propack}
Larsen, R.~M.
\newblock 2004.
\newblock Propack-software for large and sparse svd calculations.
\newblock {\em Available online. URL http://sun. stanford. edu/rmunk/PROPACK}
  2008--2009.

\bibitem[\protect\citeauthoryear{Li \bgroup et al\mbox.\egroup
  }{2016}]{li2016multitask}
Li, T.; Cheng, B.; Ni, B.; Liu, G.; and Yan, S.
\newblock 2016.
\newblock Multitask low-rank affinity graph for image segmentation and image
  annotation.
\newblock {\em ACM Transactions on Intelligent Systems and Technology (TIST)}
  7(4):65.

\bibitem[\protect\citeauthoryear{Liu \bgroup et al\mbox.\egroup
  }{2013}]{liu2013robust}
Liu, G.; Lin, Z.; Yan, S.; Sun, J.; Yu, Y.; and Ma, Y.
\newblock 2013.
\newblock Robust recovery of subspace structures by low-rank representation.
\newblock {\em IEEE Transactions on Pattern Analysis and Machine Intelligence}
  35(1):171--184.

\bibitem[\protect\citeauthoryear{Maaten and
  Hinton}{2008}]{maaten2008visualizing}
Maaten, L. v.~d., and Hinton, G.
\newblock 2008.
\newblock Visualizing data using t-sne.
\newblock {\em Journal of machine learning research} 9(Nov):2579--2605.

\bibitem[\protect\citeauthoryear{Ng \bgroup et al\mbox.\egroup
  }{2002}]{ng2002spectral}
Ng, A.~Y.; Jordan, M.~I.; Weiss, Y.; et~al.
\newblock 2002.
\newblock On spectral clustering: Analysis and an algorithm.
\newblock {\em NIPS} 2:849--856.

\bibitem[\protect\citeauthoryear{Nie, Wang, and
  Huang}{2014}]{nie2014clustering}
Nie, F.; Wang, X.; and Huang, H.
\newblock 2014.
\newblock Clustering and projected clustering with adaptive neighbors.
\newblock In {\em SIGKDD},  977--986.
\newblock ACM.

\bibitem[\protect\citeauthoryear{Niyogi}{2004}]{niyogi2004locality}
Niyogi, X.
\newblock 2004.
\newblock Locality preserving projections.
\newblock In {\em NIPS}, volume~16,  153.
\newblock MIT.

\bibitem[\protect\citeauthoryear{Passalis and
  Tefas}{2017}]{passalis2017dimensionality}
Passalis, N., and Tefas, A.
\newblock 2017.
\newblock Dimensionality reduction using similarity-induced embeddings.
\newblock {\em IEEE transactions on neural networks and learning systems}.

\bibitem[\protect\citeauthoryear{Peng \bgroup et al\mbox.\egroup
  }{2016}]{peng2016deep}
Peng, X.; Xiao, S.; Feng, J.; Yau, W.-Y.; and Yi, Z.
\newblock 2016.
\newblock Deep subspace clustering with sparsity prior.
\newblock In {\em IJCAI},  1925--1931.

\bibitem[\protect\citeauthoryear{Peng \bgroup et al\mbox.\egroup
  }{2018}]{peng2018integrate}
Peng, C.; Kang, Z.; Cai, S.; and Cheng, Q.
\newblock 2018.
\newblock Integrate and conquer: Double-sided two-dimensional k-means via
  integrating of projection and manifold construction.
\newblock {\em ACM Transactions on Intelligent Systems and Technology (TIST)}
  9(5):57.

\bibitem[\protect\citeauthoryear{Peng, Cheng, and
  Cheng}{2017}]{peng2017supervised}
Peng, C.; Cheng, J.; and Cheng, Q.
\newblock 2017.
\newblock A supervised learning model for high-dimensional and large-scale
  data.
\newblock {\em ACM Transactions on Intelligent Systems and Technology (TIST)}
  8(2):30.

\bibitem[\protect\citeauthoryear{Roweis and Saul}{2000}]{roweis2000nonlinear}
Roweis, S.~T., and Saul, L.~K.
\newblock 2000.
\newblock Nonlinear dimensionality reduction by locally linear embedding.
\newblock {\em Science} 290(5500):2323--2326.

\bibitem[\protect\citeauthoryear{Tao \bgroup et al\mbox.\egroup
  }{2017}]{taoensemble2017}
Tao, Z.; Liu, H.; Li, S.; Ding, Z.; and Fu, Y.
\newblock 2017.
\newblock From ensemble clustering to multi-view clustering.
\newblock In {\em IJCAI},  2843--2849.

\bibitem[\protect\citeauthoryear{Tenenbaum, De~Silva, and
  Langford}{2000}]{tenenbaum2000global}
Tenenbaum, J.~B.; De~Silva, V.; and Langford, J.~C.
\newblock 2000.
\newblock A global geometric framework for nonlinear dimensionality reduction.
\newblock {\em Science} 290(5500):2319--2323.

\bibitem[\protect\citeauthoryear{Towne, Ros{\'e}, and
  Herbsleb}{2016}]{towne2016measuring}
Towne, W.~B.; Ros{\'e}, C.~P.; and Herbsleb, J.~D.
\newblock 2016.
\newblock Measuring similarity similarly: Lda and human perception.
\newblock {\em ACM Transactions on Intelligent Systems and Technology} 8(1).

\bibitem[\protect\citeauthoryear{Weinberger, Blitzer, and
  Saul}{2005}]{weinberger2005distance}
Weinberger, K.~Q.; Blitzer, J.; and Saul, L.~K.
\newblock 2005.
\newblock Distance metric learning for large margin nearest neighbor
  classification.
\newblock In {\em NIPS},  1473--1480.

\bibitem[\protect\citeauthoryear{Wright \bgroup et al\mbox.\egroup
  }{2009}]{wright2009robust}
Wright, J.; Yang, A.~Y.; Ganesh, A.; Sastry, S.~S.; and Ma, Y.
\newblock 2009.
\newblock Robust face recognition via sparse representation.
\newblock {\em IEEE transactions on pattern analysis and machine intelligence}
  31(2):210--227.

\bibitem[\protect\citeauthoryear{Xu \bgroup et al\mbox.\egroup
  }{2009}]{xu2009extended}
Xu, Z.; Jin, R.; King, I.; and Lyu, M.
\newblock 2009.
\newblock An extended level method for efficient multiple kernel learning.
\newblock In {\em NIPS},  1825--1832.

\bibitem[\protect\citeauthoryear{Xu \bgroup et al\mbox.\egroup
  }{2010}]{xu2010simple}
Xu, Z.; Jin, R.; Yang, H.; King, I.; and Lyu, M.~R.
\newblock 2010.
\newblock Simple and efficient multiple kernel learning by group lasso.
\newblock In {\em ICML-10},  1175--1182.
\newblock Citeseer.

\bibitem[\protect\citeauthoryear{Zhang \bgroup et al\mbox.\egroup
  }{2017}]{zhang2017adaptive}
Zhang, L.; Zhang, Q.; Du, B.; You, J.; and Tao, D.
\newblock 2017.
\newblock Adaptive manifold regularized matrix factorization for data
  clustering.
\newblock In {\em IJCAI},  33999--3405.

\bibitem[\protect\citeauthoryear{Zhang, Yang, and Feng}{2011}]{zhang2011sparse}
Zhang, L.; Yang, M.; and Feng, X.
\newblock 2011.
\newblock Sparse representation or collaborative representation: Which helps
  face recognition?
\newblock In {\em ICCV},  471--478.
\newblock IEEE.

\bibitem[\protect\citeauthoryear{Zhao \bgroup et al\mbox.\egroup
  }{2013}]{zhao2013similarity}
Zhao, Z.; Wang, L.; Liu, H.; and Ye, J.
\newblock 2013.
\newblock On similarity preserving feature selection.
\newblock {\em IEEE Transactions on Knowledge and Data Engineering}
  25(3):619--632.

\bibitem[\protect\citeauthoryear{Zhuang \bgroup et al\mbox.\egroup
  }{2017}]{zhuang2017label}
Zhuang, L.; Zhou, Z.; Gao, S.; Yin, J.; Lin, Z.; and Ma, Y.
\newblock 2017.
\newblock Label information guided graph construction for semi-supervised
  learning.
\newblock {\em IEEE Transactions on Image Processing} 26(9):4182--4192.

\end{thebibliography}
\end{document}